\documentclass[runningheads]{llncs}
\usepackage{wrapfig}
\usepackage{graphicx}

\usepackage{amsmath,amssymb} 
\usepackage{color}
\usepackage{overpic}

\usepackage{hyperref}
\usepackage{pgfplots,tikz}

\usepackage{multirow}

\usepackage{microtype}
\newcommand{\X}{\mathcal{X}}
\newcommand{\Y}{\mathcal{Y}}

\definecolor{poseA}{rgb}{0.871,0,0.871}
\definecolor{styleA}{rgb}{1,0.6,0.4}
\definecolor{poseB}{rgb}{0.8,0.8,0}
\definecolor{styleB}{rgb}{0.184,0.333,0.592}

\usepackage{rotating}

\begin{document}

\pagestyle{headings}
\mainmatter
\def\ECCVSubNumber{4866}  

\title{LIMP: Learning Latent Shape Representations with Metric Preservation Priors}

\titlerunning{LIMP: Latent Interpolation with Metric Priors}

\author{Luca Cosmo\inst{1,2}, Antonio Norelli\inst{1}, Oshri Halimi\inst{3}, Ron Kimmel\inst{3},\\ Emanuele Rodol\`a\inst{1}}
\authorrunning{L. Cosmo et al.}

\institute{Sapienza University of Rome, Italy
\and University of Lugano, Switzerland
\and Technion - Israel Institute of Technology, Israel}

\maketitle

\begin{abstract}
In this paper, we advocate the adoption of metric preservation as a powerful prior for learning latent representations of deformable 3D shapes. 
Key to our construction is the introduction of a geometric distortion criterion, defined directly on the decoded shapes, translating the preservation of the metric on the decoding to the formation of linear paths in the underlying latent space. Our rationale lies in the observation that training samples alone are often insufficient to endow generative models with high fidelity, motivating the need for large training datasets. In contrast, metric preservation provides a rigorous way to control the amount of geometric distortion incurring in the construction of the latent space, leading in turn to synthetic samples of higher quality. We further demonstrate, for the first time, the adoption of differentiable intrinsic distances in the backpropagation of a geodesic loss. Our geometric priors are particularly relevant in the presence of scarce training data, where learning any meaningful latent structure can be especially challenging. The effectiveness and potential of our generative model is showcased in applications of style transfer, content generation, and shape completion.

\end{abstract}

\section{Introduction}

Constructing high-fidelity generative models for 3D shapes is a challenging problem that has met with increasing interest in recent years. 
Generative models are applicable in many practical domains, ranging from content creation to shape exploration, as well as in 3D reconstruction.
As a new generation of methods, they come to face a number of difficulties.
 
Most existing approaches address the case of static or {\em rigid} geometry, for example, man-made objects like chairs and airplanes, with potentially high intra-class variability; see the ShapeNet~\cite{shapenet2015} repository for such examples. 
In this setting, the main focus has been on the abstraction capabilities of the encoder and the generator, describing complex 3D models in terms of their core geometric features via parsimonious part-based representations. 
Shapes generated with these techniques are usually designed to have valid part semantics that are easy to parse. Concurrently, several recent efforts have concentrated on the definition of convenient representations for the 3D {\em output}; these methods find broader application in multiple tasks, where they enable more efficient and high-quality synthesis, and can be often plugged into existing generative models.

To date, relatively fewer approaches have targeted the {\em deformable} setting, where the generated shapes are related by continuous, non-rigid deformations. 
These model a range of natural phenomena, such as changes in pose and facial expressions of human subjects, articulations, garment folding, and molecular flexibility to name but a few. 
The extra difficulties brought by such non-rigid deformations can be tackled, in some cases, by designing mathematical or parametric models for the deformation at hand; however, these models are often violated in practice, and can be very hard to devise for general deformations -- hence the need for learning from examples. 

\begin{figure}[t]
    \centering
    	\begin{overpic}
		[trim=0cm 1cm 0cm 0cm,clip,width=0.99\linewidth]{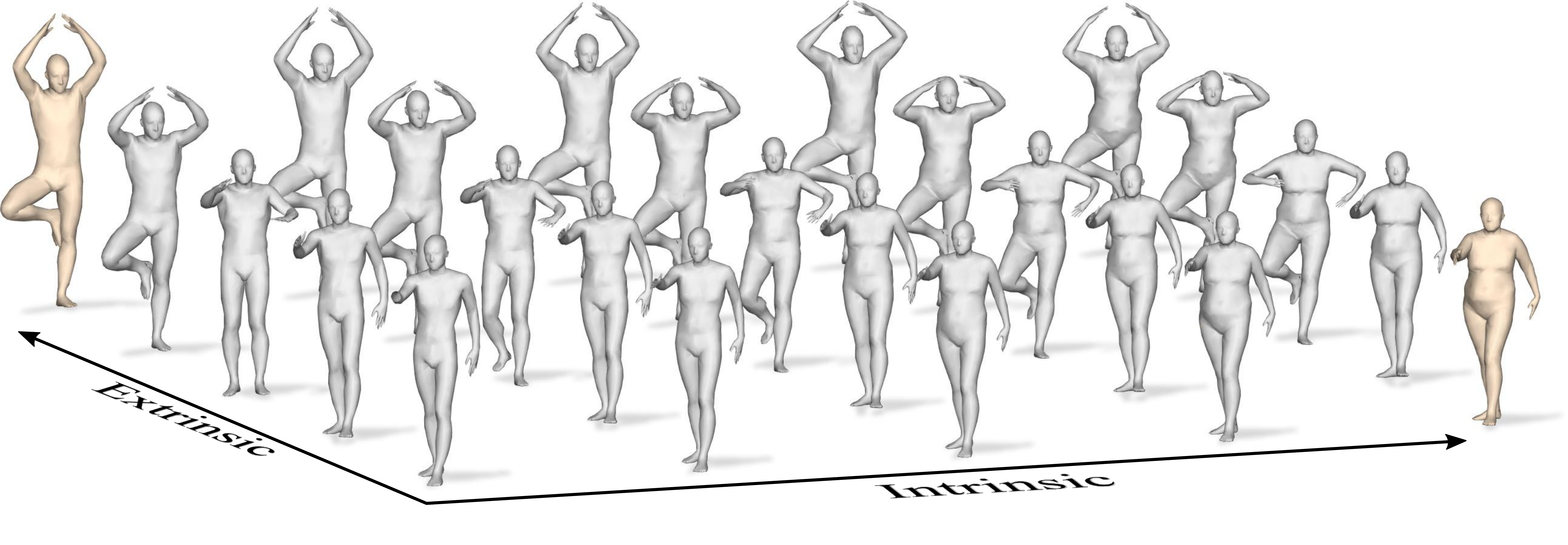}
		\end{overpic}
    \caption{Disentangled interpolation of FAUST shapes, obtained with our generative model trained under metric preservation priors. The yellow shapes at the two corners are given as input; the remaining shapes are generated by bilinearly interpolating the latent codes of the input, and decoding the resulting codes. Our model allows to disentangle pose from identity, illustrated here as different dimensions.}
    \label{fig:teaser}
\end{figure}

The framework we propose is motivated by the observation that existing data-driven approaches for learning deformable 3D shapes, and autoencoders (AE) in particular, do not make use of any {\em geometric prior} to drive the construction of the latent space, whereas they rely almost completely on the expressivity of the training dataset. 
This imposes a heavy burden on the learning process, and further requires large annotated datasets that can be costly or even impossible to acquire.
In the absence of additional regularization, limited training data leads to limited generalization capability, which is manifested in the generated 3D shapes exhibiting unnatural distortions.
Variational autoencoders (VAE) provide a partial remedy by modeling a distributional prior on the data via a parametrized density on the latent space. 
This induces additional regularization, but is still insufficient to guarantee the preservation of geometric properties in the output 3D models.

In this paper, we introduce \textit{Latent Interpolation with Metric Priors} (LIMP).
We propose to {explicitly} model the local {\em metric} properties of the latent space by enforcing metric constraints on the decoded output. 
We do this by phrasing a metric distortion penalty that has the effect to promote naturally looking deformations, and in turn to significantly reduce the need for large datasets at training time. 
In particular, we show that by coupling the Euclidean distances among latent codes (hence, along linear paths in the latent space) to the metric distortion among decoded shapes, we obtain a strong regularizing effect in the construction of the latent space.
Another novel ingredient of the proposed approach is the backpropagation of intrinsic (namely, geodesic) distances during training, which is made possible by a recent geodesic computation technique. 
Using geodesics makes our approach more flexible, and enables the successful application of our generative model to style and pose transfer applications.
See Figure~\ref{fig:teaser} for an example of novel samples synthesized with our generative model.

\section{Related Work}
Our method falls within the class of AE-based generative models for 3D shapes. In this Section we cover methods from this family that are more closely related to ours, and refer to the recent survey \cite{eg19tut} for a broader coverage.

In the 3D computer vision and graphics realms, generative models for part-compositional 3D objects play the lion's share. Such approaches directly exploit the hierarchical, structural nature of 3D man-made objects to drive the construction of encoder and generator \cite{nash2017shape,li2017grass,mo2019structurenet,nash2020polygen}. These methods leverage on the insight that objects can be understood through their components \cite{liu2018physical}, making an interpretable representation close to human parsing possible. In this setting, a continuous exploration of the generated latent spaces is not always meaningful; the mechanism underlying typical operations like sampling and interpolation happen instead in {\em discrete} steps in order to generate plausible intermediate shape configurations (e.g., for transitioning from a 4-legged chair to a 3-legged stool). 
For this reason, with rigid geometry one usually deals with ``structural blending'' rather than continuous deformations.
Structural blending has been realized, for instance, by learning abstractions of symmetry hierarchies via spatial arrangements of oriented bounding boxes~\cite{li2017grass}, or by explicitly modeling part-to-part relationships~\cite{mo2019structurenet}; generative-adversarial modeling has been applied on volumetric object representations~\cite{wu2016learning}; structural hierarchies have been applied for the generation of composite 3D scenes \cite{li2019grains} and building typologies \cite{deepform} as well. Contributing to their success, is the fact that all these methods train on ShapeNet-scale annotated datasets with $>50K$ unique 3D models, and the recent publication of dedicated benchmarks like PartNet~\cite{mo2019partnet} testify to the increasing interest of data-driven models for structure-aware geometry processing.
In this paper, we address a different setting; we do not assume part-compositionality of the 3D models since we deal with deformable shapes, where continuous deformations are well-defined, and where annotated datasets are not as prominent.

A second thread of research revolves around the definition of a meaningful representation for the generated 3D output. While many approaches mostly use polygonal meshes with predefined topology or directly synthesize point clouds \cite{achlioptas2018learning,shu20193d}, the focus has been recently shifting towards more effective representations in terms of overall quality, fidelity, and flexibility. These include approaches that predict implicit shape representations at the output, requiring an ex-post isosurface extraction step to generate a mesh at the desired resolution \cite{mescheder2019occupancy,park2019deepsdf,gropp2020implicit}; isosurfacing has been replaced by binary space partitioning in \cite{chen2019bsp}; while in \cite{groueix2018papier}, shapes are represented by a set of parametric surface elements. 
In this work, we focus on learning a better {\em latent} representation for deformable shapes, rather than on constructing a better representation for the output.

More closely related to ours are some recent methods from the area of geometric deep learning. A graph-convolutional VAE with dynamic filtering convolutional layers \cite{verma2018feastnet} was introduced in \cite{litany2018deformable} for the task of deformable shape completion of human shapes. The method is trained on $\sim$7000 shapes from the DFAUST dataset of real human scans \cite{dfaust:CVPR:2017}; due to the lack of any geometric prior, the learned generator introduces large distortions around points in the latent space that are not well represented in the training set. Geometric regularization was injected in \cite{groueix20183d} in the form of a template that parametrizes the surface. The method shows excellent performance in shape matching, however, it crucially relies on a large and representative dataset of $230,000$ shapes, and performance drops significantly with smaller training sets or bad initialization. More recently, a geometric disentanglement model for deformable point clouds was introduced in~\cite{aumentado2019geometric}. The proposed method uses Laplacian eigenvalues as a weak geometric prior to promote the separation of intrinsic and extrinsic shape information, together with several other de-correlation penalties, and a training set of $>40K$ shapes. In the absence of enough training examples, the approach tends to produce a ``morphing'' effect between point clouds that does not correspond to a natural motion; a similar phenomenon was observed in~\cite{achlioptas2018learning}.
Finally, in \cite{tan2020realtime}, a time-dependent physical prior was used to regularize interpolations in the latent space with the goal of obtaining a convincing simulation of moving tissues.

 In particular, our approach bears some analogies with the theory of shape spaces \cite{heeren2014exploring}, in that we seek to synthesize geometry that minimizes a deformation energy. For example, in~\cite{devir2009reconstruction} it was shown how to {\em axiomatically} modify a noisy shape such that its intrinsic measures would fit a given prior in a different pose. Differentiating the geodesic distances was done by fixing the update order in the fast marching scheme~\cite{FMM}. Our energy is not minimized over a fixed shape space, but rather, it drives the construction of a novel shape space in a data-driven fashion.

In this paper, we leverage classical ideas from shape analysis and metric geometry to ensure that shapes on the learned latent space correspond to plausible (i.e., low-distortion) deformations of the shapes seen at training time, even when only few training samples are available. We do this by modeling a geometric prior that promotes deformations with bounded distortion, and show that this model provides a powerful regularization for shapes {\em within} as well as {\em across} different classes, e.g., when transitioning between different human subjects.

\section{Learning with metric priors}
Our goal is to learn a latent representation for deformable 3D shapes. We do this by training a VAE on a training set $\mathcal{S}=\{\mathbf{X}_i\}$ of $|\mathcal{S}|$ shapes, under a purely geometric loss:
\begin{align}\label{eq:loss}
    \ell(\mathcal{S}) = \ell_\textrm{recon}(\mathcal{S}) + \ell_\textrm{interp}(\mathcal{S}) + \ell_\textrm{disent}(\mathcal{S})\,.
\end{align}
The loss is composed of three terms. The first is a geometric reconstruction loss on the individual training shapes, as in classical AE's; the second one is a pairwise interpolation term for points in the latent space; the third one is a disentanglement term to separate intrinsic from extrinsic information.

The main novelty lies in (1) the interpolation loss, and (2) the disentanglement loss {\em not} relying upon corresponding poses in the training set. The interpolation term provides control over the encoding of each shape {\em in relation to the others}. This induces a notion of proximity between latent codes that is explicitly linked, in the definition of the loss, to a notion of metric distortion between the decoded shapes. As we show in the following, this induces a strong regularization on the latent space and rules out highly distorted reconstructions.

The disentanglement loss promotes the factorization of the latent space into two orthogonal components: One that spans the space of isometries (e.g., change in pose), and another that spans the space of non-isometric deformations (e.g., change in identity). As in the interpolation loss, for the disentanglement we also exploit the metric properties of the decoded shapes.

\subsection{Losses}
We define $\mathbf{z}:=\mathrm{enc}(\mathbf{X})$ to be the latent code for shape $\mathbf{X}$, and $\mathbf{X}':=\mathrm{dec}(\mathbf{z})$ to be the corresponding decoding. During training, the decoder ($\mathrm{dec}$) and encoder ($\mathrm{enc}$) are updated so as to minimize the overall loss of Eq.~\eqref{eq:loss}; see Section~\ref{sec:impl} for the implementation details.

\paragraph*{Geometric reconstruction.}
The reconstruction loss is defined as follows:
\begin{align}\label{eq:recon}
    \ell_\mathrm{recon}(\mathcal{S}) = \sum_{i=1}^{|\mathcal{S}|} \| \mathbf{D}_{\mathbb{R}^3}( \mathbf{X}'_i)  - \mathbf{D}_{\mathbb{R}^3}(\mathbf{X}_i) \|_F^2\,,
\end{align}
where $\mathbf{D}_{\mathbb{R}^3}(\mathbf{X})$ is the matrix of pairwise Euclidean distances between all points in $\mathbf{X}$, and $\|\cdot\|_F$ denotes the Frobenius norm. Eq.~\eqref{eq:recon} measures the cumulative reconstruction error (up to a global rotation) over the training shapes.

\paragraph*{Metric interpolation.}
This loss is defined over all possible pairs of shapes $(\mathbf{X}_i,\mathbf{X}_j)$:
\begin{align}\label{eq:interp}
    \ell_\mathrm{interp}(\mathcal{S}) = \sum_{i\neq j}^{|\mathcal{S}|} \|
    \mathbf{D}(\mathrm{dec}(
    \underbrace{(1-\alpha)\mathbf{z}_i  + \alpha \mathbf{z}_j}_{\substack{\textrm{interpolation of}\\\textrm{latent codes}}}
    )) - 
    (
    \underbrace{(1-\alpha)\mathbf{D}( \mathbf{X}_i' ) + \alpha \mathbf{D}( \mathbf{X}_j')}_{\substack{\textrm{interpolation of}\\\textrm{geodesic or local distances}}}
    ) \|_F^2\,,
\end{align}
where $\alpha\sim \mathcal{U}(0,1)$ is a uniformly sampled scalar in $(0,1)$, different for each pair of shapes. In the equation above, the matrix $\mathbf{D}(\mathbf{X})$ encodes the pairwise distances between points in $\mathbf{X}$. We use two different definitions of distance, giving rise to two different losses which we sum up together. In one loss, $\mathbf{D}$ contains {\em geodesic} distances between {\em all} pairs of points. In the second loss, we consider {\em local Euclidean} distances from each point to points within a small neighborhood (set to $10\%$ of the shape diameter); the rationale is that local Euclidean distances capture local detail and tend to be resilient to non-rigid deformations, as observed for instance in~\cite{rodola2017partial}.
All distances are computed on the fly, on the decoded shapes, at each forward step. 


Since the error criterion in Eq.~\eqref{eq:interp} encodes the discrepancy between pairwise distance matrices, we refer to it as a {\em metric preservation prior}. We refer to Section~\ref{sec:analysis} for a more in-depth discussion from a continuous perspective.

\paragraph*{Disentanglement.}
We split the latent codes into an intrinsic and an extrinsic part, $\mathbf{z}:=(\mathbf{z}^\mathrm{int} | \mathbf{z}^\mathrm{ext})$. The former is used to encode ``style'', i.e., the space of non-isometric deformations; the latter is responsible for changes in pose, and is therefore constrained to model the space of possible isometries.

The loss is composed of two terms:
\begin{align}
\ell_\mathrm{disent}(\mathcal{S}) &= \ell_\mathrm{int} (\mathcal{S}) + \ell_\mathrm{ext} (\mathcal{S}) \,, \quad \textrm{with}\\
    \ell_\mathrm{int}(\mathcal{S}) &= \sum_{\substack{i\neq j\\\mathrm{iso}}}^{|\mathcal{S}|} \| \mathbf{D}_{\mathbb{R}^3} ( \mathrm{dec} (  \underbrace{(1-\alpha)\mathbf{z}_i^\mathrm{int} + \alpha\mathbf{z}_j^\mathrm{int}}_{\substack{\textrm{interpolation of}\\\textrm{style}}}  | \mathbf{z}_i^\mathrm{ext})) - \mathbf{D}_{\mathbb{R}^3}(\mathbf{X}_i) \|_F^2 \label{eq:lint}\\
    \ell_\mathrm{ext}(\mathcal{S}) &= \sum_{\substack{i\neq j\\\textrm{non-iso}}}^{|\mathcal{S}|} \| \mathbf{D}_g ( \mathrm{dec} ( \mathbf{z}_i^\mathrm{int} |  \underbrace{(1-\alpha)\mathbf{z}_i^\mathrm{ext} + \alpha\mathbf{z}_j^\mathrm{ext}}_{\substack{\textrm{interpolation of}\\\textrm{pose}}} )) - \mathbf{D}_g(\mathbf{X}_i) \|_F^2 \label{eq:lext}
\end{align}
The $\ell_\mathrm{int}$ term is evaluated only on isometric pairs (i.e., just a change in pose), for which we expect $\mathbf{z}_i^\mathrm{int}=\mathbf{z}_j^\mathrm{int}$. For a pair $(\mathbf{X}_i,\mathbf{X}_j)$, it requires that $\mathbf{X}_i$ can be reconstructed exactly even when its intrinsic part $\mathbf{z}_i^\mathrm{int}$ is interpolated with that of $\mathbf{X}_j$. This enforces $\mathbf{z}_i^\mathrm{int}=\mathbf{z}_j^\mathrm{int}$, thus all the pose-related information is forced to move to $\mathbf{z}^\mathrm{ext}$.

The $\ell_\mathrm{ext}$ term is instead evaluated on {\em non}-isometric pairs. Here we require that the {\em geodesic} distances of $\mathbf{X}_i$ are left untouched when we interpolate its pose with that of $\mathbf{X}_j$. This way, we force all the style-related information to be moved to $\mathbf{z}^\mathrm{int}$. We see that by having direct access to the metric on the decoded shapes, we can phrase the disentanglement easily in terms of distances.

The assumption that the metric is nearly preserved under pose changes is widely used in many shape analysis applications such as shape retrieval \cite{reuter2006laplace}, matching \cite{rodola2012game,cosmo2016matching,cosmo2017consistent,halimi2019unsupervised} and reconstruction \cite{boscaini2015shape,cosmo2019isospectralization}

\paragraph*{Relative error.}
In practice, we always measure the error on the Euclidean distances (appearing in Eqs.~\eqref{eq:recon},\eqref{eq:interp},\eqref{eq:lint}) in a {\em relative} sense. Let $\mathbf{A}$ be the ``ground truth'' Euclidean distance matrix computed on the input shape, and let $\mathbf{B}$ be its predicted reconstruction. Instead of taking  $\|\mathbf{A}-\mathbf{B}\|_F^2=\sum_{ij} (\mathbf{A}_{ij}-\mathbf{B}_{ij})^2$, we compute the relative error $\sum_{ij} \frac{(\mathbf{A}_{ij}-\mathbf{B}_{ij})^2}{\mathbf{A}_{ij}^2}$.
In our experiments, this resulted in better reconstruction of local details than by using the simple Frobenius norm.

\begin{figure}[t]
    \centering
    	\begin{overpic}
		[trim=0cm 1.6cm 0cm 0cm,clip,width=0.70\linewidth]{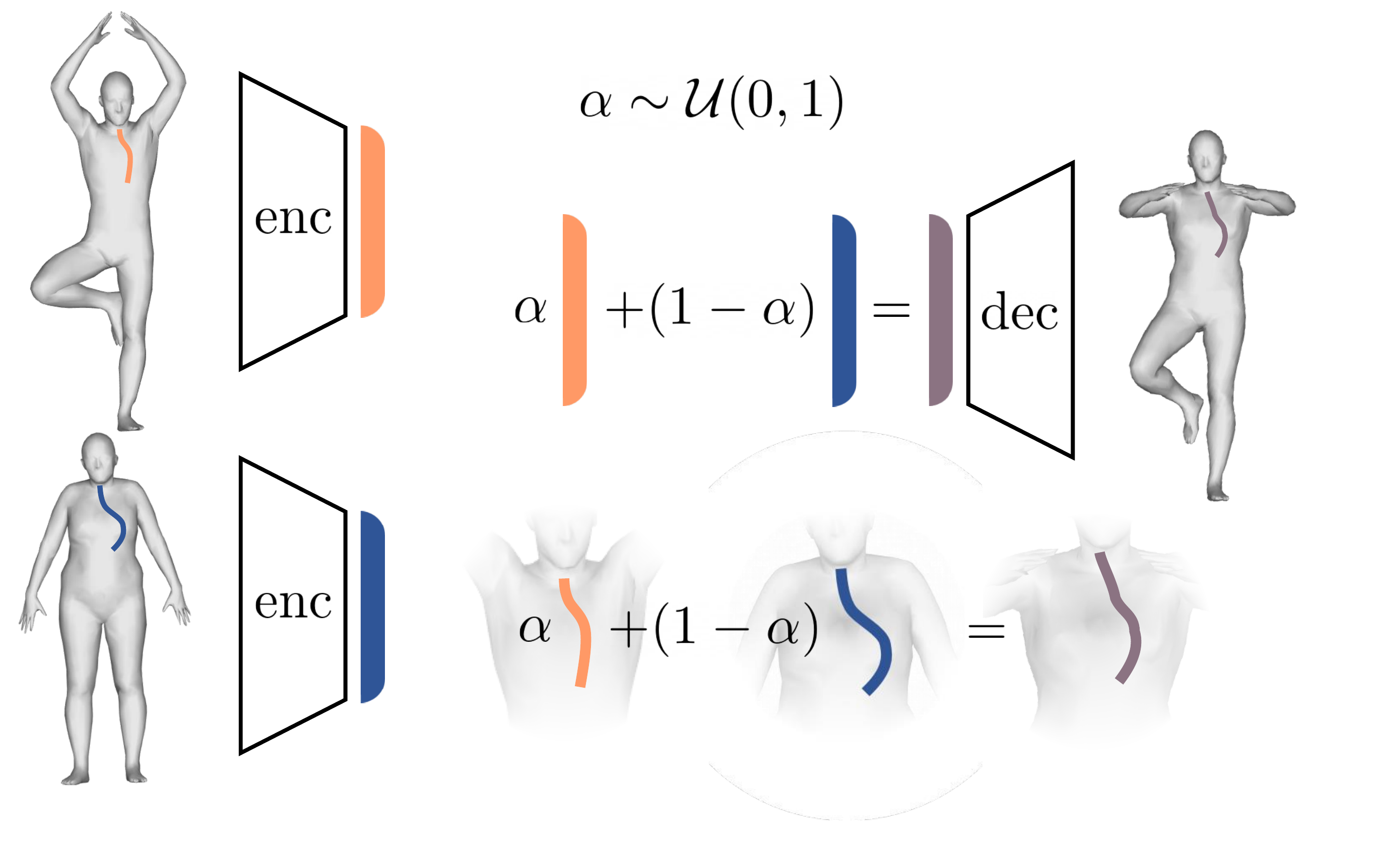}
		\end{overpic}
    \caption{\label{fig:vae}Our architecture is a standard VAE, with PointNet as the encoder and a fully connected decoder. Our loss asks that the geodesic distances on the decoded convex combination of latent codes (middle row) are equal to the the convex combination of the input distances.}
\end{figure}

\subsection{Continuous interpretation}\label{sec:analysis}
In the continuous setting, we regard shapes as metric spaces $(\X,d_\X)$, each equipped with a distance function $d_\X:\X\times\X\to\mathbb{R}_+$. Given two shapes $(\X,d_\X)$ and $(\Y,d_\Y)$, a map $\phi:\X\to\Y$ is an {\em isometry} if it is surjective and preserves distances, $d_\X(x,x')=d_\Y(\phi(x),\phi(x'))$ for all $x,x'\in\X$. Isometries play a fundamental role in 3D shape analysis, since they provide a mathematical model for natural deformations like changes in pose. In practice, however, isometry is rarely satisfied exactly. 

\paragraph*{Why interpolation?}
Our approach is based on the insight that {\em non}-isometric shapes are related by sequences of near-isometric deformations, which, in turn, have a well defined mathematical model.
In our setting, we do not require the training shapes to be near-isometric. Instead, we allow for maps $\phi$ with {\em bounded metric distortion}, i.e., for which there exists a constant $K>0$ such that:
\begin{align}\label{eq:lip}
    |d_\X(x,x')-d_\Y(\phi(x),\phi(x')) | \le K
\end{align}
for all $x, x'\in\X$. 
For $K\to 0$ the map $\phi$ is a near-isometry, while for general $K>0$ we get a much wider class of deformations, going well beyond simple changes in pose. We therefore assume that there exists a map with bounded distortion between all shape pairs in the training set.

At training time, we are given a map $\phi:\X\to\Y$ between two training shapes $(\X,d_\X)$ and $(\Y,d_\Y)$. We then assume there exists an abstract metric space $(\mathcal{L}, d_\mathcal{L})$ where each point is a shape; this ``shape space'' is the latent space that we seek to represent when training our generative model.
Over the latent space we construct a parametric sequence of shapes $\mathcal{Z}_\alpha = (\X,d_\alpha)$, parametrized by $\alpha\in(0,1)$, connecting $(\X,d_\X)$ to $(\Y,d_\Y)$. By modeling the intermediate shapes as $(\X,d_\alpha)$, we regard each $\mathcal{Z}_\alpha$ as a continuously deformed version of $\X$, with a different metric defined by the interpolation:
\begin{align}\label{eq:intd}
    d_\alpha(x,x') = (1-\alpha)d_\X(x,x') + \alpha d_\Y(\phi(x),\phi(x'))\,,
\end{align}
for all $x,x'\in\X$. Each $\mathcal{Z}_\alpha$ in the sequence has the same points as $\X$, but the shape is different since distances are measured differently.

It is easy to see that if the training shapes $\X$ and $\Y$ are isometric, then $d_\alpha(x,x')=d_\X(x,x')$ for all $x,x'\in\X$ and the entire sequence is isometric, i.e., we are modeling a change in pose. However, if $\phi:\X\to\Y$ has bounded distortion without being an isometry, each intermediate shape $(\X,d_\alpha)$ also has bounded distortion with respect to $(\X,d_\X)$, with $K_\alpha<K$ in Eq.~\eqref{eq:lip}; in particular, for $\alpha\to 0$ one gets $K_\alpha \to 0$ and therefore a near-isometry. In other words, by using the metric interpolation loss of Eq.~\eqref{eq:interp}, as $\alpha$ grows from $0$ to $1$ we are modeling a general non-isometric deformation as a sequence of approximate isometries.

\setlength{\columnsep}{7pt}
\setlength{\intextsep}{1pt}
\begin{wrapfigure}[5]{r}{0.5\linewidth}
\vspace{-0.2cm}
  \begin{overpic}
		[trim=0cm 0cm 0cm 0cm,clip,width=0.9\linewidth]{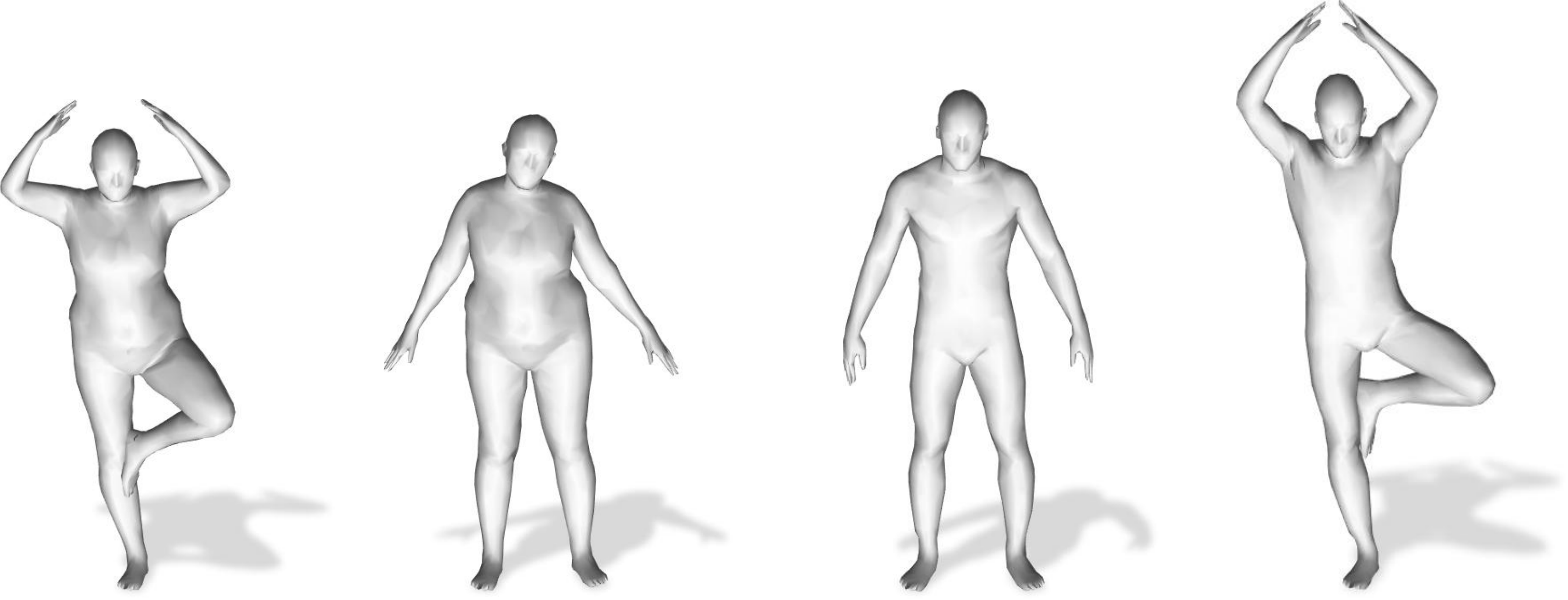}
		\put(16,15){\large $\mathbf{-}$  }
		\put(45,15){\large $\mathbf{+}$  }
		\put(74,15){\large $\mathbf{=}$  }
		\end{overpic}
\end{wrapfigure}
\paragraph*{Flattening of the latent space.}
Taking a linear convex combination of latent vectors as in Eq.~\eqref{eq:interp} implies that distances between codes should be measured using the Euclidean metric $\| \cdot \|_2$. 
This enables algebraic manipulation of the codes and the formation of ``shape analogies'', as shown in the inset (real example based on our trained model). 
By the connection of Euclidean distances in the latent space with intrinsic distances on the decoder's output, our learning model performs a ``flattening'' operation, in the sense that it requires the latent space to be as Euclidean as possible, while absorbing any embedding error in the decoder. A similar line of thought was followed, in a different context, in the purely axiomatic model of \cite{shamai2017geodesic}.

\subsection{Implementation}\label{sec:impl}
We design our deep generative model as a VAE (Figure~\ref{fig:vae}). The input data is a set of triangle meshes; each mesh is encoded as a matrix of vertex positions $\mathbf{X}\in\mathbb{R}^{n\times 3}$, together with connectivity encoded as a $n\times n$ adjacency matrix. 
We anticipate here that mesh connectivity is never accessed directly by the network.

\paragraph*{Architecture.}
The encoder takes vertex positions $\mathbf{X}$ as input, and outputs a $d$-dimensional code $\mathbf{z}=\mathrm{enc}(\mathbf{X})$. Similarly, the decoder outputs vertex positions $\mathbf{Y} = \mathrm{dec}(\mathbf{z}) \in \mathbb{R}^{n\times 3}$.
In order to clarify the role of our priors versus the sophisticacy of the architecture, we keep the latter as simple as possible.
In particular, we adopt a similar architecture as in \cite{aumentado2019geometric}; we use PointNet \cite{qi2017pointnet} with spatial transform as the encoder, and a simple MLP as the decoder. We reserve $25\%$ of the latent code for the extrinsic part and the remaining $75\%$ for the intrinsic representation, while the latent space and layer dimensions vary depending on the dataset size. A detailed description of the network is deferred to the Supplementary Material.
We implemented our model in PyTorch using Adam as optimizer with learning rate of 1e-4. To avoid local minima and numerical errors in gradient computation, we start the training by optimizing just the reconstruction loss for $10^4$ iterations, and add the remaining terms for the remaining epochs. 

\paragraph*{Geodesic distance computation.}
A crucial ingredient to our model is the computation of geodesic distances $\mathbf{D}_g(\mathrm{dec}(\mathbf{z}))$ during training, see Eq.~\eqref{eq:interp}. We use the heat method of \cite{crane13} to compute these distances, based on the realization that its pipeline is fully differentiable. It consists, in particular, of two linear solves and one normalization step, and all the quantities involved in the three steps depend smoothly on the vertex positions given by the decoder (we refer to the Supplementary Material for additional details).

To our knowledge, this is the first time that on-the-fly computation of geodesic distances appears in a deep learning pipeline.
Previous approaches using geodesic distances, such as \cite{halimi2019unsupervised}, do so by taking them as pre-computed input data, and leave them untouched for the entire training procedure. 

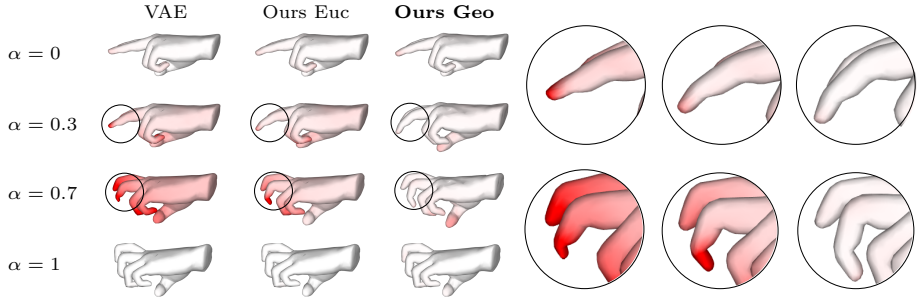
\begin{figure}[t]
    \centering
    	\begin{overpic}
		[trim=-9cm 1.4cm 0cm 0cm,clip,width=0.99\linewidth]{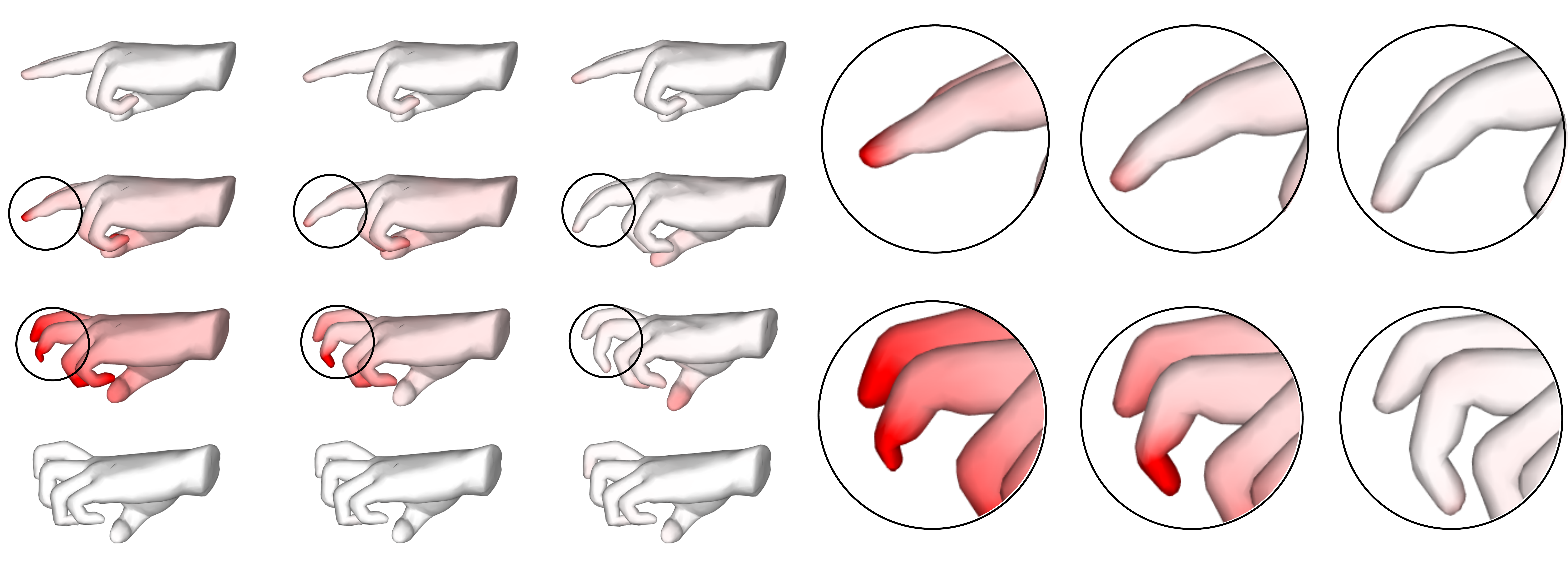}
		\put(0,3.2){\scriptsize $\alpha=1$  }
		\put(0,11){\scriptsize $\alpha=0.7$  }
		\put(0,18.7){\scriptsize $\alpha=0.3$  }
		\put(0,26.5){\scriptsize $\alpha=0$  }
		\put(15,31){\scriptsize VAE}
		\put(28,31){\scriptsize Ours Euc}
		\put(42.5,31){\scriptsize \textbf{Ours Geo}}
		\end{overpic}
    \caption{Interpolation example on a small training set of just 5 shapes, where the deformation evolves from top ($\alpha=0$) to bottom ($\alpha=1$). Color encodes the per-point metric distortion, growing from white to red; changes in pose as in this example should have distortion close to zero. We show the results obtained by three different networks: baseline VAE; ours with Euclidean metric regularization only; ours with Euclidean {\em and} geodesic regularization (i.e., the complete loss).}
    \label{fig:hands}
\end{figure}

\begin{table}[b]
\caption{\label{tab:ablation}Ablation study in terms of interpolation and disentanglement error on 4 datasets. Our full pipeline (denoted by `Ours Geo') achieves the minimum error in all cases, and is \textbf{more than one order of magnitude} better than the baseline VAE on the interpolation. We do not report the disentanglement error for HANDS, since the dataset only contains one hand style.}

\def\-{\scalebox{0.3}[1.0]{\( - \)}}
\setlength{\tabcolsep}{6pt}
\centering
\begin{tabular}{l|ccc|ccc}
\multicolumn{1}{c}{} & \multicolumn{3}{c}{Interpolation Error} & \multicolumn{3}{c}{Disentanglement Error}\\
        & VAE & Ours Euc  & Ours Geo & VAE & Ours Euc  & Ours Geo\\ 
\hline
FAUST   &  $3.89e\-2$ & $5.08e\-3$ & $\mathbf{3.82e\-3}$ &  $7.16$ & $4.04$ & $\mathbf{3.48}$  \\
DFAUST  &  9.82e-2 & 3.43e-3 & $\mathbf{2.89e\-4}$ &  $6.15$ & $4.90$ & $\mathbf{4.11}$ \\
COMA    &  $1.32e\-3$ & $1.03e\-3$ & $\mathbf{7.51e\-4}$ &  $1.55$ & $1.30$ & $\mathbf{1.22}$ \\
HANDS   &  $6.01e\-3$ & $8.12e\-4$ & $\mathbf{4.62e\-4}$ &  - & - & - \\
\end{tabular}

\end{table}

\paragraph*{Supervision.} 
We train on a collection of shapes with known pointwise correspondences; these are needed in  Eq.~\eqref{eq:interp}, where we assume that the distance matrices have compatible rows and columns. From a continuous perspective, we need maps for the interpolated metric of Eq.~\eqref{eq:intd} to be well defined.
Known correspondences are also needed by other approaches dealing with deformable data \cite{litany2017deep,litany2018deformable,groueix20183d}. In practice, we only need few such examples (we use $<100$ training shapes), since we rely for the most part on the regularization power of our geometric priors.

Differently from \cite{litany2018deformable,groueix20183d} we do {\em not} assume the training shapes to have the same mesh, since the latter is only used as an auxiliary structure for computing geodesics in the loss; the network only ever accesses vertex positions. Further, we do not require training shapes with similar poses across different subjects.

\begin{figure}[t]
    \centering
    		\begin{overpic}
		[trim=0cm 0.4cm 0cm 0cm,clip,width=1\linewidth]{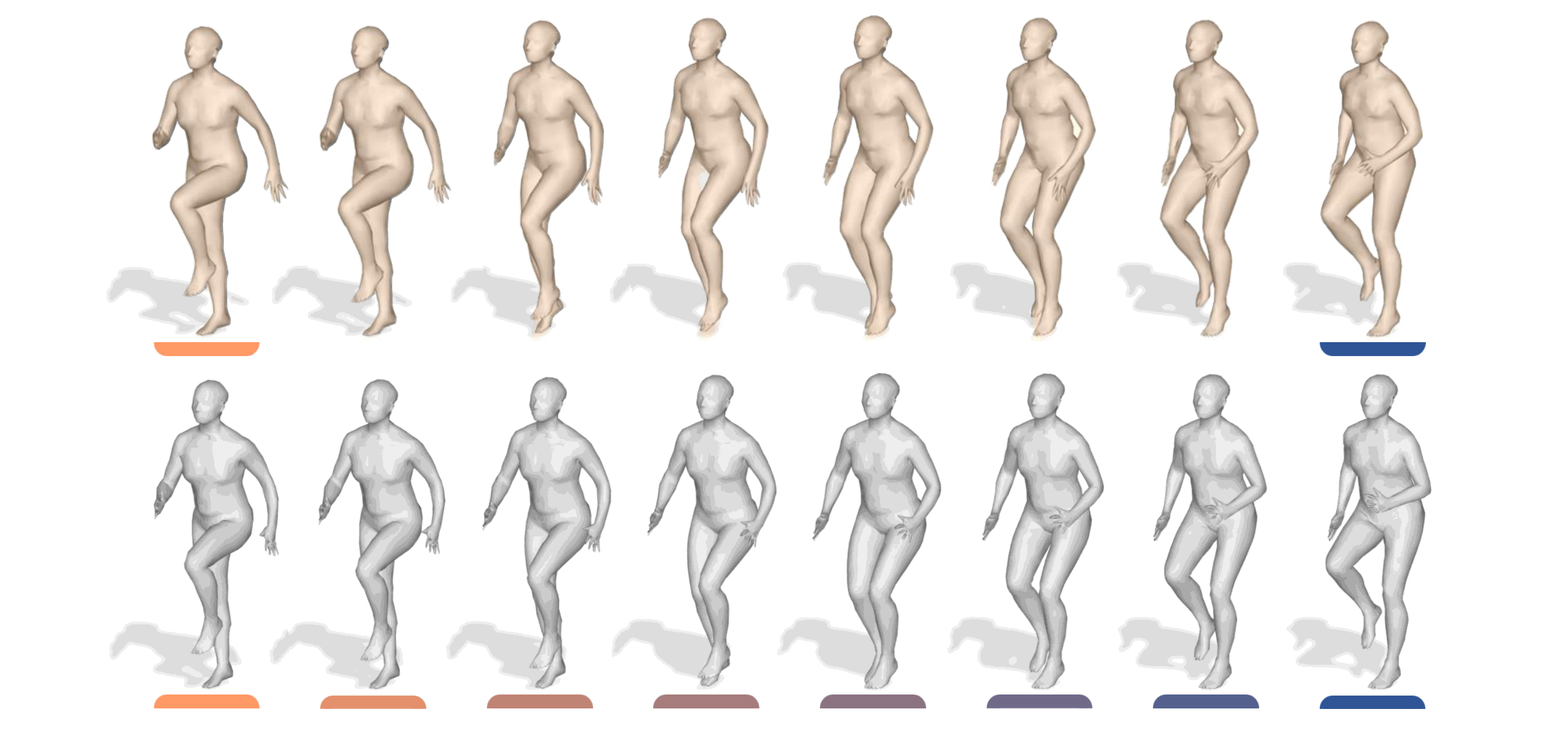}
		\end{overpic}
    \caption{\label{fig:dfaust_seq}{\em Top row}: A 4D sequence from the real-world dataset DFAUST. We train our generative model on the left- and right-most keyframes (indicated by the orange and blue bar respectively), together with keyframes extracted from other sequences and different individuals. {\em Bottom row}: The 3D shapes generated by our trained model. Visually, both the generated and the real-world sequences look plausible, indicating that geometric priors are well-suited for regularizing toward realistic deformations.}
\end{figure}

\begin{figure}[t] 
    \centering
    		\begin{overpic}
		[trim=0cm -1cm -1cm -1cm,clip,width=1\linewidth]{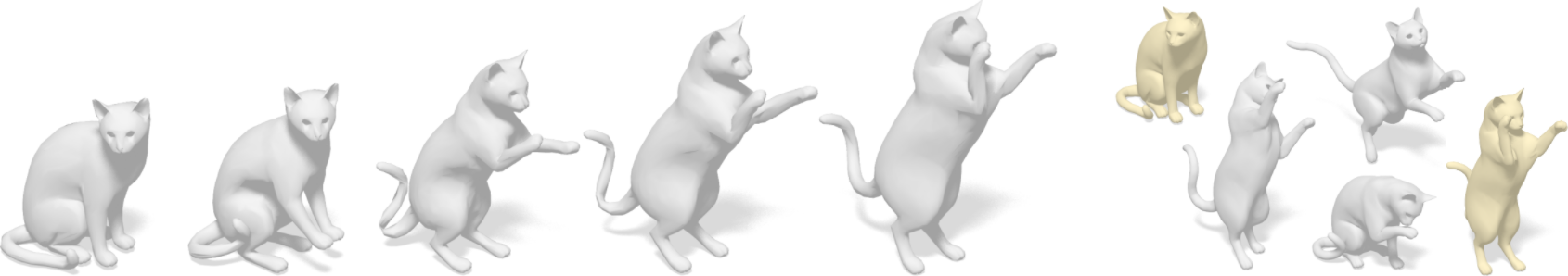}
		\put(73,18.5){\tiny Most similar training shapes}
		
		\put(70,20){\line(1,0){30}}
		\put(70,0){\line(1,0){30}}
		
		\put(70,0){\line(0,1){20}}
		\put(100,0){\line(0,1){20}}

		\end{overpic}
    \caption{Interpolation example on the \textit{cat} shapes of TOSCA dataset \cite{bronstein2008numerical}. On the left, we show an interpolation sequence between two shapes of the training set (yellow shapes on the right). On the right, we manually selected the most similar shapes present in the training set, composed in total by just 11 shapes. You can appreciate how shapes in the middle of the interpolated sequence significantly differ from the training shapes.
    }
    \label{fig:tosca}
\end{figure}
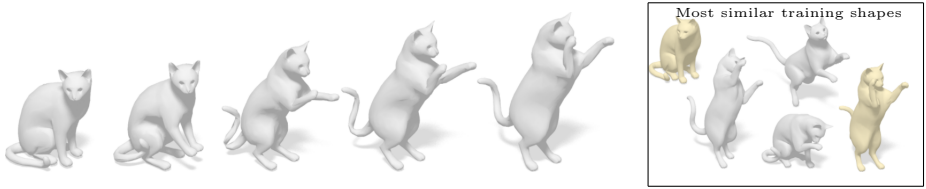

\section{Results}

\subsection{Data}
%
%
To validate our method, we performed experiments using 5 different datasets (3 are obtained from real-world scans, 2 are fully synthetic). 
\textbf{FAUST} \cite{Bogo:CVPR:2014} is composed of 10 different human subjects, each captured in 10 different poses. We train our network on 8 subjects (thus, 80 meshes in total) and leave out the other 2 subjects for testing.
\textbf{DFAUST} \cite{dfaust:CVPR:2017} is a 4D dataset capturing the motion of 10 human subjects performing 14 different activities, spanning {\em hundreds} of frames each. As training data \textbf{we only use 4 representative frames from each subject/sequence pair}.
\textbf{COMA} \cite{COMA:ECCV18} is another 4D dataset of human faces; it is composed of 13 subjects, each performing 13 different facial expressions represented as a sequence of 3D meshes. As opposed to the test split proposed in \cite{ranjan2018generating}, where 90\% of the data is used for training, we only select 14 frames for each subject (one representative for each of the 13 expressions, plus one in a neutral pose), thus \textbf{training with less than 1\% of the dataset}. 
\textbf{TOSCA} \cite{bronstein2008numerical} is a synthetic dataset containing both animals and human bodies. In our experiments we use only the \textit{cat} class, containing 11 shapes in different poses.
The last dataset, which we refer to as \textbf{HANDS}, is also completely synthetic and consists of 5 meshes depicting one hand in 5 different poses.
For all the datasets, we subsample the meshes to 2500 vertices by iterative edge collapse \cite{garland1997surface}.

\subsection{Interpolation}
We first perform a classical interpolation experiment. Given two shapes $\mathbf{X}$ and $\mathbf{Y}$, we visualize the decoded interpolation of their latent codes, given by $\mathrm{dec}((1-\alpha)\mathrm{enc}(\mathbf{X})+\alpha\mathrm{enc}(\mathbf{Y}))$ for a few choices of $\alpha\in (0,1)$. 
We measure the interpolation quality via the {\em interpolation error}, defined as the average (over all surface points) geodesic distortion of the interpolated shapes.

Two examples of interpolation are shown in Figures~\ref{fig:hands} and~\ref{fig:tosca}. In these examples, the training sets consist of just \textbf{5} and \textbf{11 shapes} respectively, meaning that the intermediate poses have never been seen before. In this few-shot setting, proper regularization is crucial to get meaningful results.
In the experiment in Figure~\ref{fig:hands}, we also conduct an ablation study. We disable all the interpolation terms from our complete loss, resulting in a baseline VAE; then we disable the geodesic regularization only; finally we keep the entire loss intact, showing best results.
Quantitative results on 4 different datasets are reported in Table~\ref{tab:ablation} (first 3 columns), showing that best results are obtained when our full loss is used.

\begin{figure}[t]
    \centering
    	\begin{overpic}
		[trim=0cm 0.7cm 0cm 0cm,clip,width=1\linewidth]{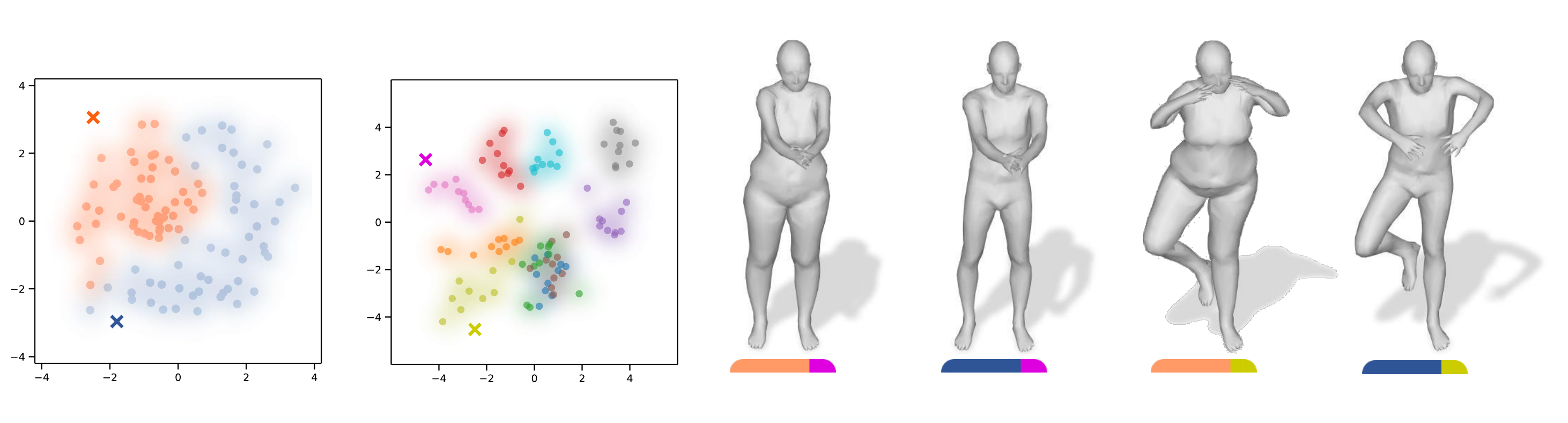}
		\put(0,20.5){\scriptsize Intrinsic latent space}
		\put(23,20.5){\scriptsize Extrinsic latent space}
		\end{overpic}
    \caption{\label{fig:factors}{\em Plots on the left}: Planar embedding of the intrinsic and extrinsic parts of the latent codes from FAUST. Colors identify gender (left) and pose (ten different poses; right). We observe cohesive clusters in either case, suggesting that the encoder has generalized the projection onto each factor. The four small crosses are random samples. {\em Right}: Decoded shapes from the four combinations of the random samples; the specific combinations are illustrated by compatible colors between the crosses and the bars below each shape.}
\end{figure}

As an additional qualitative experiment, in Figure~\ref{fig:dfaust_seq} we show the decoded shapes in-between two keyframes of a 4D sequence from DFAUST. We remark that none of the intermediate shapes were seen at training time, nor was any similar-looking shape present in the training set. We then compare our reconstructed sequence with the original sequence of real-world scans. The purpose of this experiment is to show that our geometric priors are essential for the generation of realistic motion; apart from a perceptual evaluation, any quantitative comparison here would not be meaningful -- there is not a unique ``true'' way to transition between two given poses.

\subsection{Disentanglement}
Our second set of experiments is aimed at demonstrating the effectiveness of our geometric priors for the disentanglement of intrinsic from extrinsic information. We illustrate this in different ways.

In Figure~\ref{fig:factors}, we show disentanglement for a generator trained on the FAUST dataset. For visualization purposes, for each vector $\mathbf{z}:=(\mathbf{z}^\mathrm{int}|\mathbf{z}^\mathrm{ext})$ in the latent space (here comprising both training and test shapes), we embed the $\mathbf{z}^\mathrm{int}$ and $\mathbf{z}^\mathrm{ext}$ parts {\em separately} onto the plane (via multidimensional scaling), and attribute different colors to different gender and poses. We then randomly sample two new $\mathbf{z}^\mathrm{int}$ and two new $\mathbf{z}^\mathrm{ext}$, and compose them into four latent codes by taking all the combinations. The figure illustrates the four decoded shapes.

\begin{figure}[t]
    \centering
    		\begin{overpic}
		[trim=0cm 0cm 0cm 0cm,clip,width=0.9\linewidth]{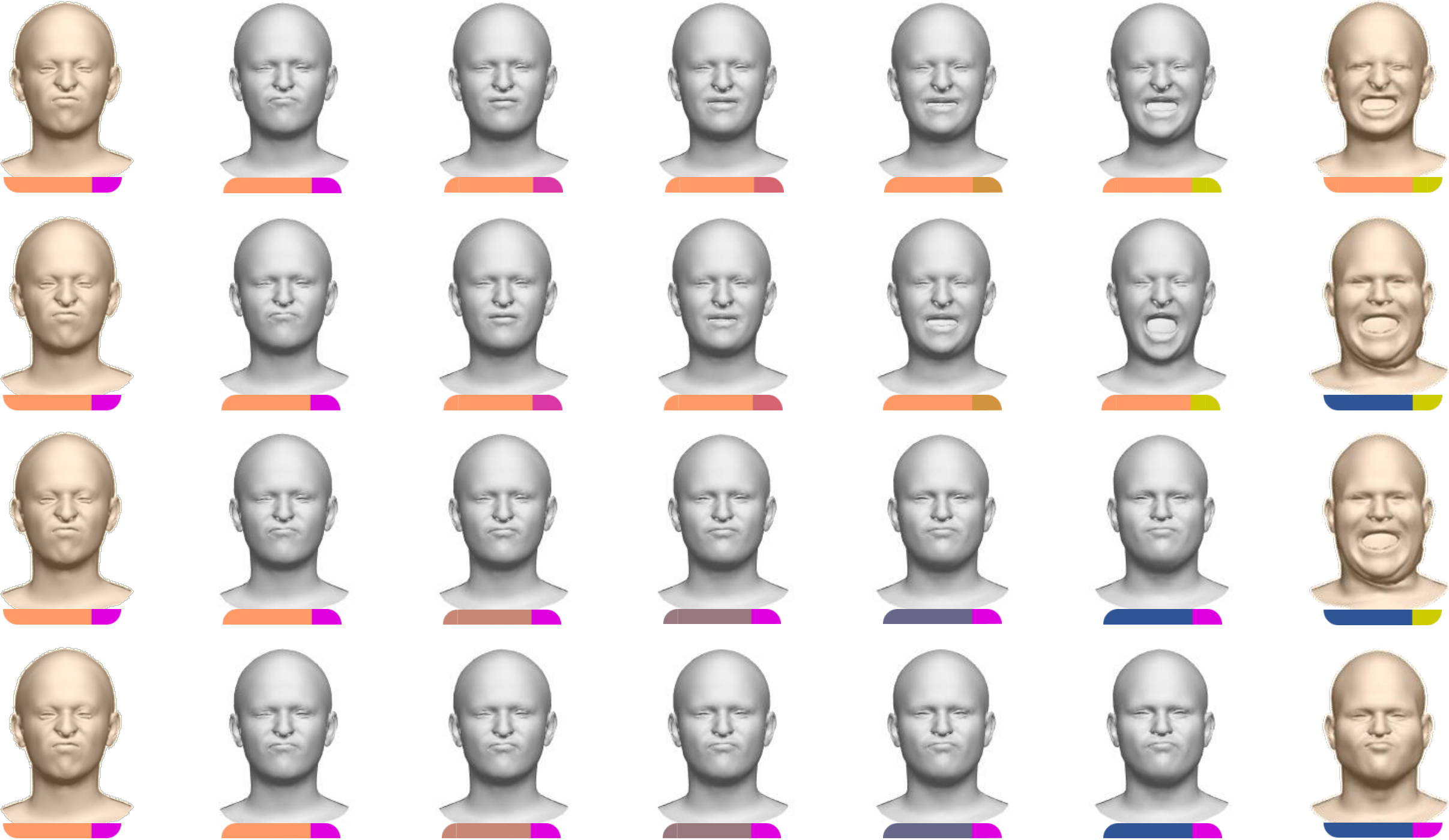}
		\put(-5,37){\begin{turn}{90}\tiny \textbf{fix style}\end{turn}}
		\put(-3,35){\begin{turn}{90}\tiny \textbf{interp. pose}\end{turn}}
		\put(-5,7){\begin{turn}{90}\tiny \textbf{interp. style}\end{turn}}
		\put(-3,9){\begin{turn}{90}\tiny \textbf{fix pose}\end{turn}}
		\end{overpic}
    \caption{Disentanglement $+$ interpolation examples on the COMA dataset; the source shape is always the same. Each row presents a different scenario, with interpolation happening left-to-right. Please refer to the color code below each shape as a visual aid; for example, for the first column we have $({\color{styleA}\textrm{\textbf{style}}}|{\color{poseA}\textrm{\textbf{pose}}})$.}
    \label{fig:discoma}
\end{figure}
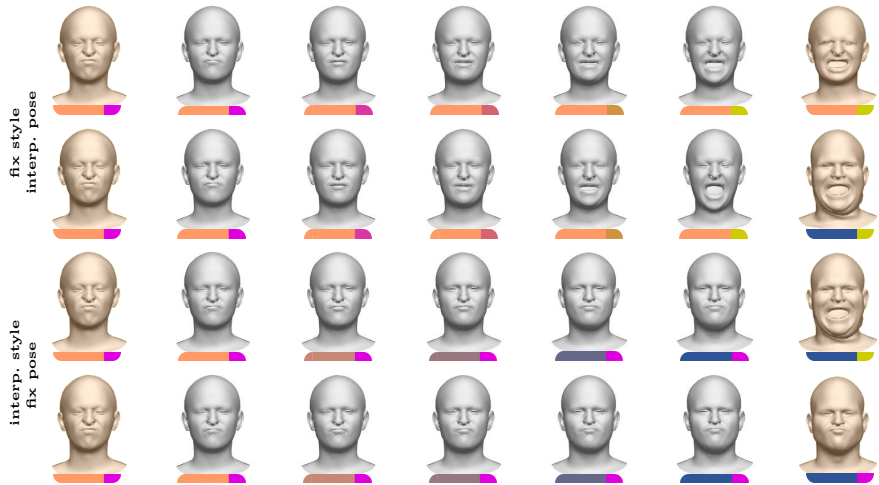

In Figure~\ref{fig:discoma} we show the simultaneous action of disentanglement and interpolation. Given a source and a target shape, we show the interpolation of pose while fixing the style, and the interpolation of style while fixing the pose. We do so with different combinations of source and target. In all cases, our generative model is able to synthesize realistic shapes with the correct semantics, suggesting high potential in style and pose transfer applications.

As we did with the case of interpolation, we also provide a notion of {\em disentanglement error}, defined as follows. Given shapes $\mathbf{X}_i$ and $\mathbf{X}_j$ with latent codes $({\color{styleA}\mathbf{z}^\mathrm{int}_i} | {\color{poseA}\mathbf{z}^\mathrm{ext}_i})$ and $({\color{styleB}\mathbf{z}^\mathrm{int}_j} | {\color{poseB}\mathbf{z}^\mathrm{ext}_j})$, we swap ${\color{poseA}\mathbf{z}^\mathrm{ext}_i}$ with ${\color{poseB}\mathbf{z}^\mathrm{ext}_j}$ and then measure the average point-to-point distance between $\mathrm{dec}({\color{styleA}\mathbf{z}^\mathrm{int}_i} | {\color{poseB}\mathbf{z}^\mathrm{ext}_j})$ and the corresponding ground-truth shape from the dataset.
In Table~\ref{tab:ablation} (last 3 columns) we report the disentanglement error on all 4 datasets, together with the ablation study.

\begin{figure}[t]
    \centering
    	\begin{overpic}
		[trim=0cm 1cm 0cm 0cm,clip,width=0.75\linewidth]{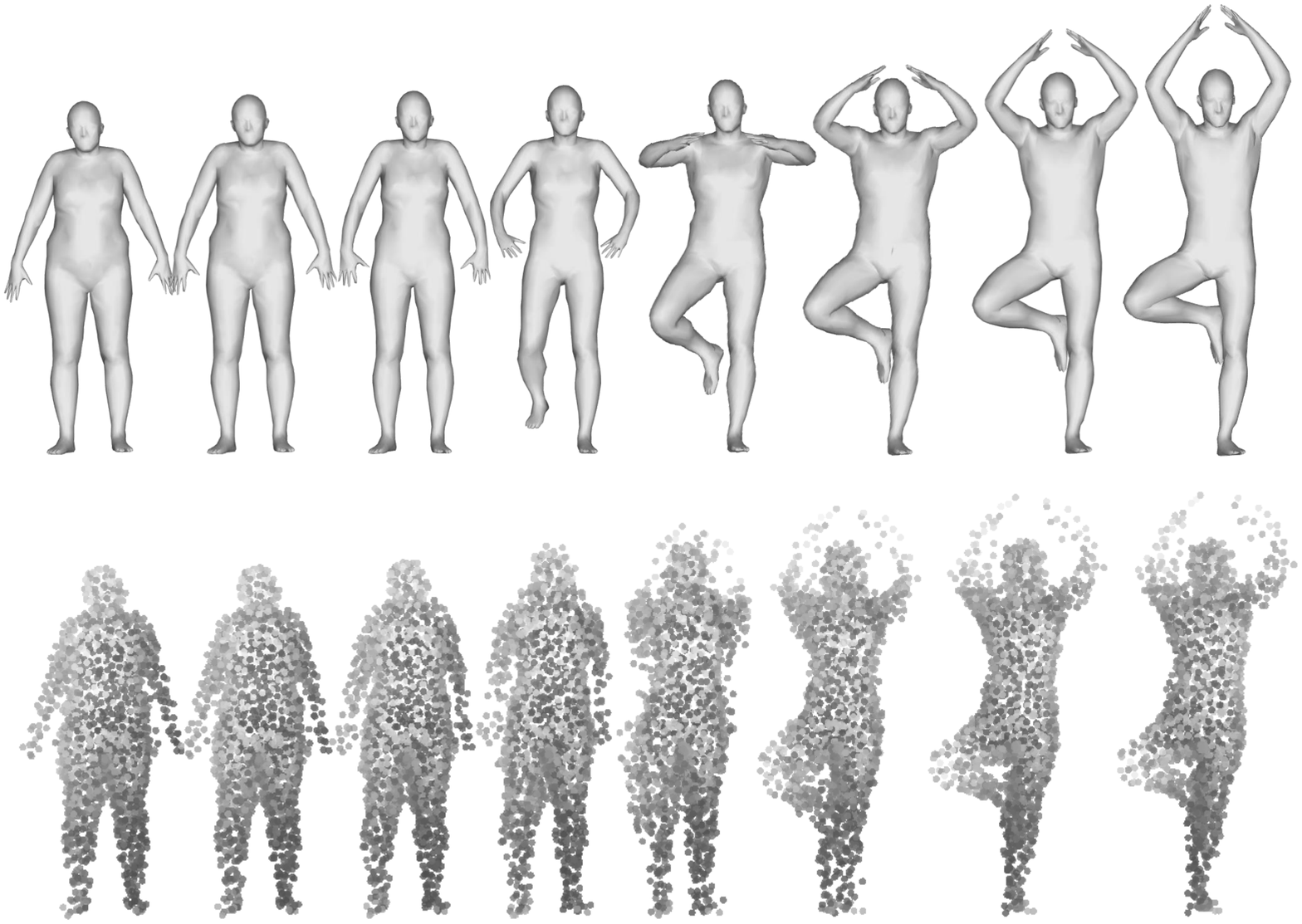}
		\end{overpic}
    \caption{\label{fig:tristan}Comparison of our method (top row) with the state-of-the-art method of \cite{aumentado2019geometric} (bottom row). Both generative models are trained on the same data. The leftmost and rightmost shapes are from the training set, while the intermediate shapes are decodings of a linear sequence in the latent space. Observe that source and target are {\em not} isometric; according to our continuous interpretation of Sec.~\ref{sec:analysis}, our trained model decomposes the non-isometric deformation into a sequence of approximate isometries.}
\end{figure}

Finally, in Figure~\ref{fig:tristan} we show a qualitative comparison with the recent state-of-the-art method~\cite{aumentado2019geometric} (using public code provided by the authors), which uses Laplacian eigenvalues as a prior to drive the disentanglement, together with multiple other de-correlation terms. Similarly to other approaches like \cite{litany2018deformable,tan2018variational}, the quality of the interpolation of \cite{aumentado2019geometric} mostly depends on the smoothness properties of the VAE, on the complexity of the deep net, or on the availability of vast training data. For this comparison, both generative models were trained on the same 80 FAUST shapes.

\section{Conclusions}
We introduced a new deep generative model for deformable 3D shapes. Our model is based on the intuition that by directly connecting the Euclidean distortion of latent codes to the metric distortion of the decoded shapes, one gets a powerful regularizer that induces a well-behaved structure on the latent space. Our idea finds a theoretical interpretation in modeling deformations with bounded metric distortion as sequences of approximate isometries.
Under the manifold hypothesis, our metric preservation priors explicitly promote a flattening of the true data manifold onto a lower-dimensional Euclidean representation. 
We demonstrated how having access to the metric of the decoded shapes during training enables high-quality synthesis of novel samples, with practical implications in tasks of content creation and style transfer.

Perhaps the main \textbf{limitation} of our method, which we share with other geometric deep learning approaches, lies in the requirement of labeled pointwise correspondences between the training shapes. These can be hard to obtain in certain settings, for example, when dealing with shapes from the same semantic class but with high intra-class variability. 
Few interesting directions of future work may consist in a self-supervised variant of our model, where dense correspondences are not needed for the training, but are estimated during the learning process 
or in the exploitation of spectral properties of the reconstructed shape, that has been shown \cite{cosmo2019isospectralization,rampini2019correspondence} to contain important information of the embedding geometry.

Finally, while in this paper we showed that even a simple prior such as metric distortion can have a significant effect, we foresee that bringing techniques from the areas of shape optimization and analysis closer to deep generative models will enable a fruitful line of stimulating research. 

\subsubsection*{Acknowledgments.}
LC, AN and ER are supported by the ERC Starting Grant No. 802554 (SPECGEO) and the MIUR under grant ``Dipartimenti di eccellenza 2018-2022'' of the Department of Computer Science of Sapienza University. OH and RK are supported by the Israel Ministry of Science and Technology grant number 3-14719, the Technion Hiroshi Fujiwara Cyber Security Research Center and the Israel Cyber Directorate.

\bibliographystyle{splncs04}
\bibliography{egbib}

\end{document}


\pagestyle{headings}
\mainmatter
\def\ECCVSubNumber{4866}  

\title{$\ast$SUPPLEMENTARY$\ast$\\LIMP: Learning Latent Shape Representations with Metric Preservation Priors}



\titlerunning{LIMP Supplementary Material}
\author{Luca Cosmo\inst{1,2}, Antonio Norelli\inst{1}, Oshri Halimi\inst{3}, Ron Kimmel\inst{3},\\ Emanuele Rodol\`a\inst{1}}
\authorrunning{L. Cosmo et al.}

\institute{Sapienza University of Rome, Italy
\and University of Lugano, Switzerland
\and Technion - Israel Institute of Technology, Israel}
\maketitle

\begin{abstract}
This supplementary material includes technical details and additional results that were excluded from the main submission due to lack of space. All the results are obtained with exactly the same methodology as described in the main manuscript. This document is also accompanied by a video, showing animated examples of disentanglement and interpolation in the latent space. We strongly encourage visualization of the accompanying video\footnote{\href{https://youtu.be/jSszaaMTg9Q}{https://youtu.be/jSszaaMTg9Q}}.
\end{abstract}


\section{Implementation details}\label{sec:supimpl}

\subsection{Architecture \& optimization}
Our architecture takes the form of a variational autoencoder (VAE) \cite{kingma2013auto} operating on point clouds. The dimension of the bottleneck changes depending on the dataset, as described below, the rationale being that different shape classes have different kinds of variability; for example, human facial expressions tend to be less articulated than animals changing their pose.

The Encoder is based on the PointNet~\cite{qi2017pointnet} architecture. The initial $x,y,z$ coordinates are given as input to a Spatial Transform (ST) module, followed by few layers of PointNet Convolution. Max-pooling throughout point features is then applied to obtain a global shape feature which is projected onto the latent space via a MLP. The output of the Encoder is actually twice the size of the latent space since it predicts both mean and variance of the multivariate Gaussian distribution from which the latent code is sampled. 

The Decoder is a simple MLP, transforming the input latent space vector to the corresponding triangular mesh embedding in $\mathbb{R}^3$.
We report in Table~\ref{tab:dim} the layer dimensions for each dataset.

\begin{table}[t]
  \caption{Detailed description of layers size used for each dataset. }
    \centering
    \begin{tabular}{l|c|c|c}
        & FAUST, DFAUST & COMA & TOSCA, HANDS\\
        \hline \hline
         \textsc{Encoder} & & \\
         ST (Conv) & 64,128,512 & - & -\\
         ST (MLP) & 512,256 & - & -\\
         \cline{2-4}
         PointNet (Conv) & 32,128,256,512 & 64, 128 & 64,64,128\\
         PointNet (MLP)  & 512,258,128,256 & 128, 64, 64 & 128,64,64,64\\
         \hline
         \textsc{Latent Space} ($\mathbf{z}^{\textrm{int}}|\mathbf{z}^{\textrm{ext}}$) & 128 (96$|$32) & 32 (24$|$8) & 32\\
         \hline
         \textsc{Decoder}  & 128,1024,2048,$N\times 3$ & 32,256,$N\times 3$ & 32,64,128,$N\times 3$\\
         
    \end{tabular}
  
    \label{tab:dim}
\end{table}

\vspace{1ex}\noindent\textbf{Optimization.}
Our loss is composed additively by three terms:
\begin{align}\label{eq:suploss}
    \ell(\mathcal{S}) = \ell_\textrm{recon}(\mathcal{S}) + \ell_\textrm{interp}(\mathcal{S}) + \ell_\textrm{disent}(\mathcal{S})\,.
\end{align}
The interpolation and disentanglement terms involve the computation of geodesic distances, which might get unstable if the underlying surface is self-intersecting or degenerate. For this reason, we disable $\ell_\textrm{interp}$ and $\ell_\textrm{disent}$ for the first $10^4$ training epochs. The action of the reconstruction loss alone provides an initial guess for the decoded shape; once this is in place, we re-activate the two missing losses and complete the training process.

\vspace{1ex}\noindent\textbf{Training times.}
One crucial point of our method is the backpropagation through all pairwise geodesic distances during training. Standard fast-marching algorithms are usually not differentiable. To overcome this limitation we use a loss based on intrinsic distances, leveraging a very efficient (and differentiable) way of calculating the geodesics, i.e. the heat method \cite{crane13}. Its computational complexity is guaranteed to be sub-quadratic in the total number of vertices , and an efficient implementation using sparse matrices has near-linear cost. In practice, our distances are computed one order of magnitude faster than a state-of-the-art fast marching implementation. Once computed, we have to keep distance values in memory. In our system, with an Intel Xeon E5-2620v4 and one Nvidia Titan Xp, it takes around 4 hours to train on FAUST dataset and around 2.67e-3 seconds for inference (2.44e-3 for encoding and 2.19e-4 for decoding). The runtime cost for the computation of the loss is dominated by the calculation of geodesic distances (80\%), which are involved in the interpolation and disentanglement terms of the loss.

\subsection{Geodesic distance calculation}
Our loss involves the calculation of geodesic distances between all points of a given decoded shape $\mathbf{X}$. In particular, the distance calculation must be differentiable with respect to the vertex positions encoded in $\mathbf{X}$. To this end, we employ the geodesics in heat algorithm introduced in~\cite{crane13}. The calculation involves three steps:
\begin{enumerate}
    \item Solve the linear system $( \mathbf{I} - t\mathbf{L}(\mathbf{X}))\mathbf{u} = \mathbf{u}_i$
    \item Compute the normalized gradient field $\mathbf{U}=-\frac{\mathbf{G(\mathbf{X})u}}{\| \mathbf{G(\mathbf{X})u} \|}$
    \item Solve the linear system $\mathbf{L}(\mathbf{X}) \mathbf{d}_i = \mathbf{D}(\mathbf{X})\mathbf{U}$
\end{enumerate}
In step (1) the time parameter is given and fixed to $t=10^{-1}$, while $\mathbf{u}_i$ is an indicator vector at point $i$. The solution vector $\mathbf{d}_i$ in step (3) contains the geodesic distance from point $i$ to all other points in $\mathbf{X}$.

The matrices $\mathbf{D}(\mathbf{X})$, $\mathbf{G}(\mathbf{X})$, and $\mathbf{L}(\mathbf{X})$ involved in these steps represent, respectively, the {\em Laplacian}, {\em gradient}, and {\em divergence} operators over a mesh with vertices $\mathbf{X}$; the coordinates in $\textbf{X}$ change at each forward step during training.

\vspace{1ex}\noindent\textbf{Remark.} This is the only point in our pipeline where a mesh structure is needed, although we remark that {\em any} mesh can be used (and in particular, different meshes for different shapes are allowed); these operators can also be defined directly on point clouds \cite[Sec. 3.2.3]{crane13}, although under some regularity condition on the point density.

\vspace{1ex}
The standard formulas that define $\mathbf{D}(\mathbf{X})$, $\mathbf{G}(\mathbf{X})$, and $\mathbf{L}(\mathbf{X})$ in terms of $\mathbf{X}$ can be found, for instance, in \cite[Sec. 3.2.1]{crane13}. Apart from some degenerate cases (which we rule out by the incremental optimization of Sec.~\ref{sec:supimpl}), all the operations involved in their definition are differentiable with respect to $\mathbf{X}$.

\section{Additional results}
We present here some additional results and comparisons that could not fit in the main manuscript due to space limitations. 

\setlength{\columnsep}{1pt}
\setlength{\intextsep}{1pt}
\begin{wrapfigure}[6]{r}{0.55\linewidth}
\vspace{-0.17cm}
\centering
  \begin{overpic}
		[trim=0cm 0cm 0cm 0cm,clip,width=0.9\linewidth]{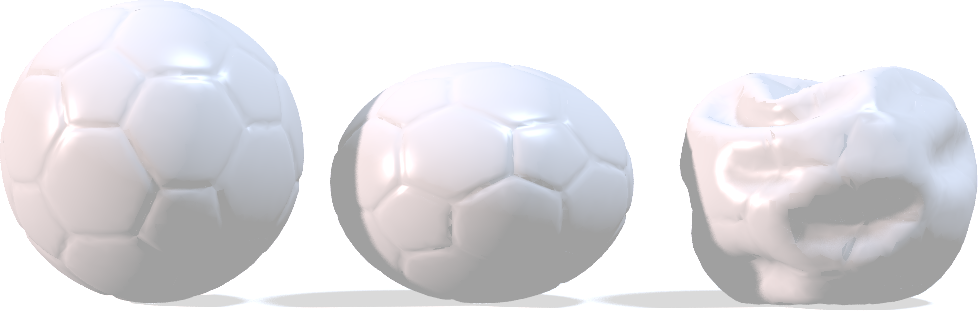}
		\put(8,-5){\footnotesize target}
		\put(7.5,-10){\footnotesize metric}
		\put(45.5,-5){\footnotesize init.}
		\put(72,-5){\footnotesize \textbf{optimized}}
		\end{overpic}
\end{wrapfigure}
\subsection{Effect of the geodesic prior}
We first illustrate how the metric preservation prior under the {\em geodesic} metric induces realistic deformations, and therefore helps in learning a well behaved generative model for deformable objects. To this end, we designed the following simple experiment. We modify the loss of Eq.~\eqref{eq:suploss} by just keeping the geometric reconstruction term; then, in this reconstruction loss, we use the {\em geodesic} distance between all points instead of the Euclidean distance. We then run the following optimization for the example shown in the inset figure:
\begin{enumerate}
    \item Compute the geodesic distances $\mathbf{D}_\textrm{tar}$ for the leftmost shape;
    \item Initialize the optimization with the middle shape;
    \item Optimize for a shape having geodesic distances $\mathbf{D}_\textrm{opt}\approx \mathbf{D}_\textrm{tar}$.
\end{enumerate}
The result is a shape having approximately the same metric as the target ball, but with a different embedding in $\mathbb{R}^3$; hence the deflated effect. Therefore, by prescribing the geodesic metric one gets a type of regularization that avoids excessive surface stretching and compression, while promoting limited distortion in the reconstruction (decoding). In the accompanying \textbf{video}, we show a comparison between the adoption of metric priors vs. no priors, demonstrating that the former lead to better synthesis of novel shapes.

\subsection{Shape completion of real scans}
One straightforward application of our generative model is  {\em deformable} shape completion. Given a partial view of a deformable shape, the task is to complete this partial view in a semantically plausible manner. For this task, we follow verbatim the pipeline of Litany et al.~\cite{litany2018deformable}, but we replace their VAE with ours. The completion procedure consists in searching over the learned latent space for a latent vector which, once decoded, produces a shape that has a close overlap with the input part. We implemented this idea, where we measured the quality of the overlap by the asymmetric Chamfer distance between part and completion.

\begin{figure}[t]
    \centering
    	\begin{overpic}
		[trim=0cm 0cm 0cm 0cm,clip,width=0.80\linewidth]{./figures/mhad}
		\put(13,35.5){\footnotesize \textbf{Ours}}
		\put(27,35.5){\footnotesize \cite{litany2018deformable}}
		\put(51,35.5){\footnotesize \textbf{Ours}}
		\put(64,35.5){\footnotesize \cite{litany2018deformable}}
		\end{overpic}
        \begin{overpic}
		[trim=0cm 0cm 0cm 0cm,clip,width=0.8\linewidth]{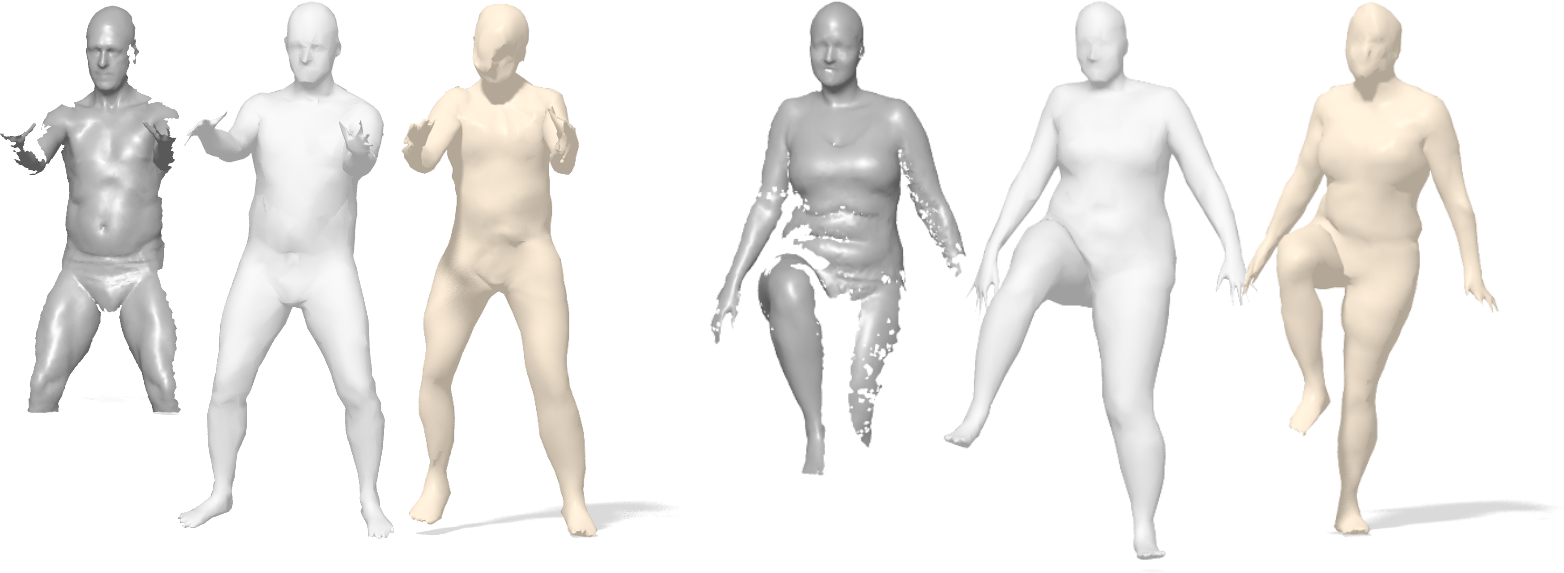}
		\put(14.5,-2.3){\footnotesize \textbf{Ours}}
		\put(30.5,-2.3){\footnotesize \cite{litany2018deformable}}
		\put(63.7,-2.3){\footnotesize \textbf{Ours}}
		\put(85,-2.3){\footnotesize \cite{litany2018deformable}}
		\end{overpic}
    \caption{\label{fig:completion}Examples of deformable shape completion obtained with our method and with the method of Litany et al.~\cite{litany2018deformable}. {\em Top row}: Completion of real Kinect scans from the MHAD dataset~\cite{mhad}; we show the original 2.5D frames on the right. {\em Bottom row}: Completion of scans from the DFAUST dataset~\cite{dfaust:CVPR:2017}.}
\end{figure}

The results are shown in Figure~\ref{fig:completion}, where we compare directly with the method of \cite{litany2018deformable} over {\em real scans} from the MHAD dataset \cite{mhad} (Kinect) and from the DFAUST dataset~\cite{dfaust:CVPR:2017}.
The training data for this experiment consists of \textbf{just 4 representative frames} for each subject/sequence pair, resulting in $<500$ training shapes in total; this is opposed to Litany et al.~\cite{litany2018deformable}, which is trained on a dataset of $\sim 7000$ shapes.

\begin{figure}[th!]
    \centering
    	\begin{overpic}
		[trim=0cm 1cm 0cm 0cm,clip,width=0.88\linewidth]{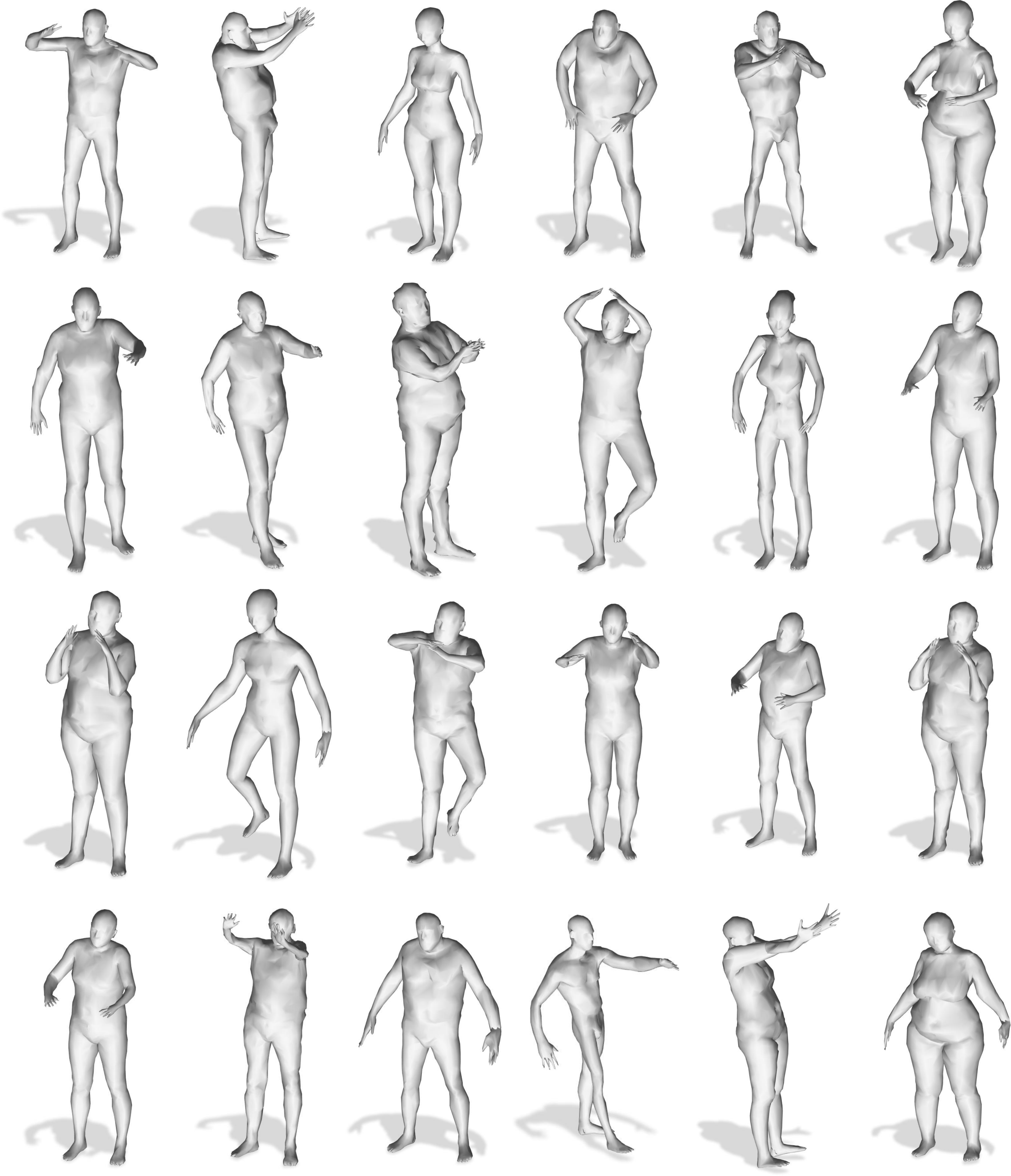}
		\end{overpic}
    \caption{\label{fig:sampling}Random samples from a latent space trained on FAUST. We sample a normal distribution with large standard deviation, thus going outside of the ``known region'' seen at training time. This generates individuals with new and sometimes extreme body features, although always within a bounded metric distortion as imposed by our priors.}
\end{figure}

\subsection{Sampling of the latent space}
In Figure~\ref{fig:sampling} we show random samples from our learned latent space, trained on 80 shapes of the FAUST dataset~\cite{Bogo:CVPR:2014}. The sampling is done according to a normal distribution with standard deviation equal to \textbf{3 times} the one of the training set, thus going well beyond the span of the known shapes, allowing us to qualitatively evaluate how much our metric priors help to regularize the generative process.

\begin{figure}[th]
    \centering
    	\begin{overpic}
		[trim=0cm 0cm 0cm 0cm,clip,width=0.93\linewidth]{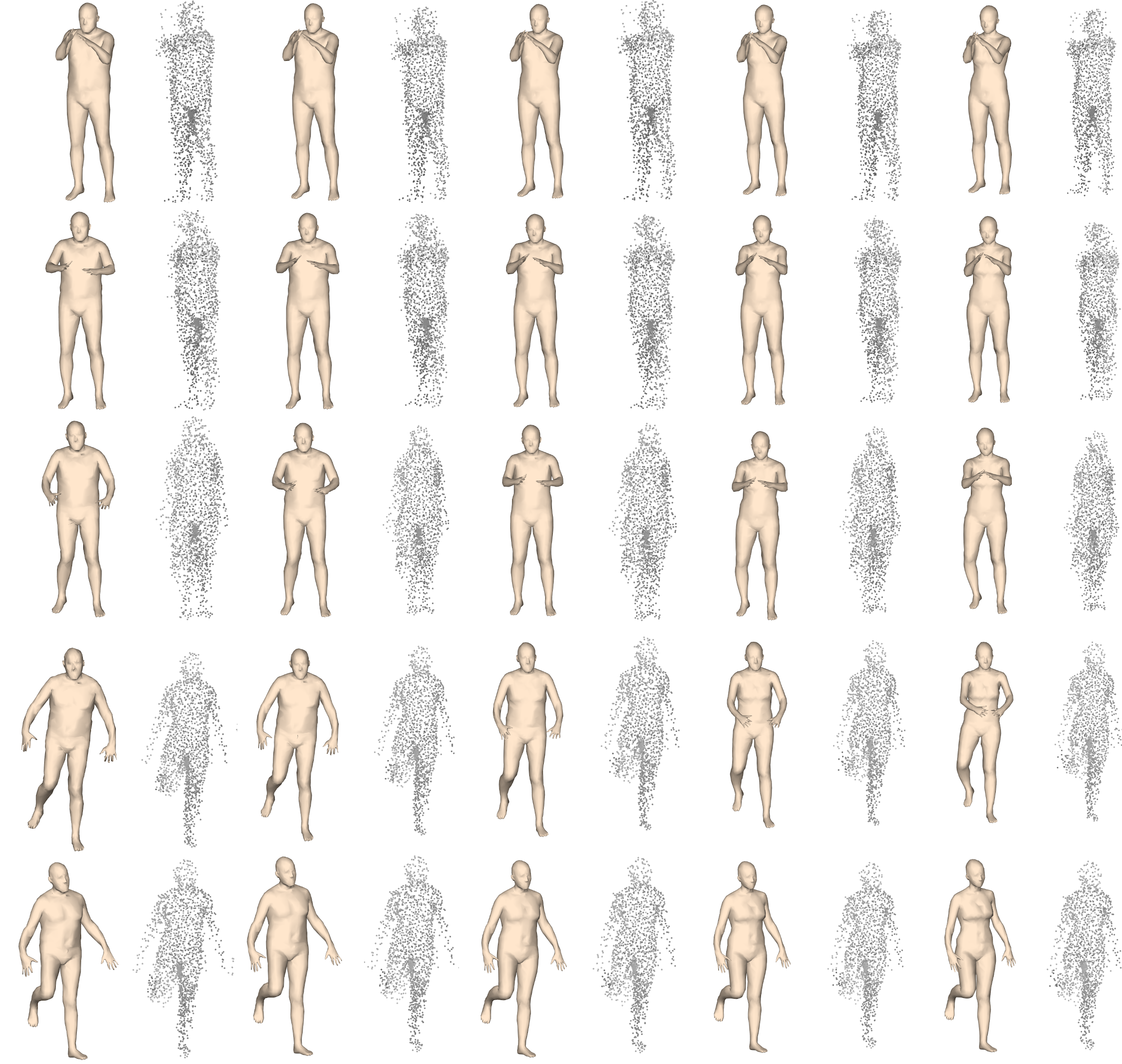}
		\end{overpic}
    \caption{\label{fig:tristan2}Comparison with the disentanglement method of \cite{aumentado2019geometric} on FAUST data. The yellow meshes are generated with our model, while the gray point clouds are obtained with \cite{aumentado2019geometric}. Style changes horizontally, while pose changes vertically.}
\end{figure}

\subsection{Comparison with the state of the art}
We show additional comparisons with the recent state-of-the-art method of Aumentado-Armstrong et al.~\cite{aumentado2019geometric}, in terms of shape interpolation and disentanglement. For this experiment, we trained both methods on the same 80 FAUST shapes. The two training shapes used as end-points for the interpolation are shown on the top-left and bottom-right of Figure~\ref{fig:tristan2}.

\bibliographystyle{splncs04}
\bibliography{egbib}